\documentclass{ceurart}

\usepackage{mathtools}
\usepackage{csquotes}
\usepackage{todonotes}
\usepackage{siunitx}
\usepackage{subcaption}

\newcommand{\mP}{\mathbb{P}}

\newcommand{\bx}{\boldsymbol{x}}
\newcommand{\KLD}{\operatorname{KL}}

\begin{document}

\copyrightyear{2022}
\copyrightclause{Copyright for this paper by its authors.
  Use permitted under Creative Commons License Attribution 4.0
  International (CC BY 4.0).}


\title{Full Kullback-Leibler-Divergence Loss for Hyperparameter-free Label Distribution Learning}

\author[1,2]{Maurice Günder}[%
orcid=0000-0001-9308-8889,
email=mguender@uni-bonn.de,
url=,
]
\address[1]{Fraunhofer-Institute for Intelligent Analysis and Information Systems (IAIS), Schloss Birlinghoven, 53757 Sankt Augustin, Germany}
\address[2]{Institute for Computer Science III, University of Bonn, Friedrich-Hirzebruch-Allee 8, 53115 Bonn, Germany}

\author[1]{Nico Piatkowski}[%
orcid=0000-0002-6334-8042,
email=nico.piatkowski@iais.fraunhofer.de,
url=,
]

\author[1,2]{Christian Bauckhage}[%
orcid=0000-0001-6615-2128,
email=christian.bauckhage@iais.fraunhofer.de,
url=,
]

\begin{abstract}
  The concept of Label Distribution Learning (LDL) is a technique to stabilize classification and regression problems with ambiguous and/or imbalanced labels. A prototypical use-case of LDL is human age estimation based on profile images. Regarding this regression problem, a so called Deep Label Distribution Learning (DLDL) method has been developed. The main idea is the joint regression of the label distribution and its expectation value. However, the original DLDL method uses loss components with different mathematical motivation and, thus, different scales, which is why the use of a hyperparameter becomes necessary. In this work, we introduce a loss function for DLDL whose components are completely defined by Kullback-Leibler (KL) divergences and, thus, are directly comparable to each other without the need of additional hyperparameters. It generalizes the concept of DLDL with regard to further use-cases, in particular for multi-dimensional or multi-scale distribution learning tasks.
\end{abstract}

\begin{keywords}
  Label distribution learning \sep
  Kullback-Leibler divergence \sep
  Machine learning
\end{keywords}

\maketitle

\section{Introduction}

Nowadays, machine learning and, particularly, deep learning techniques are used for vast amounts of classification tasks. Especially if classification problems can be formulated within an ordinal scale, a transfer into a regression problem is a favorable practice. However, the ambiguity of labels challenges modeling processes, both for classification~\cite{label_ambig_classification} and regression~\cite{ldl1} problems. In order to tackle this problem, the concept of \enquote{Label Distribution Learning} (LDL) has been developed~\cite{ldl1,dldl}. It can be interpreted as a mixture of a classification and a regression task. The overall goal is to reformulate the regression of a label with a distribution represented by a set of classifiers. This technique is shown to stabilize regression or segmentation tasks with ambiguous labels and on imbalanced data sets~\cite{ldl1, dldl}. A convenient approach is to formulate the label distribution as a probability density function (pdf) due to its mathematical properties. Maybe the most common measure for comparing the similarity of two distributions is the Kullback-Leibler (KL) divergence~\cite{kl_div}. The KL divergence is continuously differentiable and, thus, can be used as a loss function for gradient-based optimization techniques like deep learning. In this work, we propose an age estimation model based on a Convolutional Neural Network (CNN), whereas the corresponding regression problem is addressed by learning a pdf of the true age, conditioned on an input image. The general task is referred to as \enquote{Deep Label Distribution Learning} (DLDL)~\cite{dldl}. We explain shortcomings of existing DLDL loss functions and propose a more unified approach that is invariant of label scales and does not require any hyperparameters.

\section{Methodology}

The main idea of DLDL~\cite{dldl} is to reformulate a regression problem by discretizing the value range of the regression target and then learning the model parameters such that the probability mass function (pmf) over the predicted discrete class is close to a desired probability density of the true regression target.

Here, the predicted distribution is represented by the output neurons of a neural network's dense output layer---each output neuron corresponds to one discrete bin, i.e., one age value. To preserve both, the label distribution and the actual true label value, the loss is formulated as the sum of a (discrete) Kullback-Leibler (KL) divergence loss $L_{KL}$ for the label distribution $\mP(y\mid\bx)$ and an L1-loss $L_{L1}$ for the expectation value $\mu$ of this distribution. 
\begin{align}
    L_{KL} &= \KLD(\mP||\hat{\mP}) = \sum_{y} \mP(y\mid\bx)\log\frac{\mP(y\mid\bx)}{\hat{\mP}(y\mid\bx)}\,,\\\label{kl_div}
    L_{L1} &= |\mu - \hat{\mu}|\,,
\end{align}
where $y$ is the label and $\hat{\mP}(y\mid \bx)$ denotes the corresponding predicted label distribution for some data point $\bx$.
A softmax activation function~\cite{softmax} on the last layer of the network ensures that the model output $\hat{\mP}(y\mid\bx)$ is a properly normalized pmf.
The true label distribution $\mP(y\mid\bx)$ is assumed to be part of the input. In practice, we may retrieve it from problem specific expert knowledge or use some simple surrogate, e.g., the normal distribution. 

In this setup, the model is optimized jointly on $L_{KL}$ and $L_{L1}$. However, a weighting between label distribution and expectation loss is necessary, since both function have a different scaling and thus, the loss with the larger absolute value will have a stronger effect on the learning process. To overcome this issue, a hyperparameter $\lambda$ is introduced. Thus, the total loss is
\begin{align}
    L = L_{ld} + \lambda L_{exp} \coloneqq L_{KL} + \lambda L_{L1}\,,
\end{align}
where we rephrase $L_{KL}$ as the label distribution loss $L_{ld}$ and $L_{L1}$ as the expectation loss $L_{exp}$ in order to put emphasis on their functions in the label distribution learning process.

Nevertheless, when it comes to multi-dimensional labels with different scales and units, a hyperparameter is hard to choose and tuning itself is very time consuming. Thus, we propose to eliminate $\lambda$ by introducing a loss, where all components are defined in a KL fashion. The advantage is, that the mathematical motivation behind all components is the same which aligns their scales and domains automatically.

\subsection{The New Holistic KL Loss}
When rephrasing all loss components as a Kullback Leibler divergence, the $L_{ld}$ term will keep its form. For $L_{exp}$, we introduce a KL loss defined as the KL divergence between true and predicted distributions as if they were normal distributions
\begin{align}
    \mathcal{N}(y\mid\mu,\sigma^2) = \frac{1}{\sqrt{2\pi\sigma^2}} \exp\left(-\frac{(y-\mu)^2}{2\sigma^2}\right)
\end{align}
with expectation value $\mu$ and variance $\sigma^2$. $\mu$ and $\sigma^2$ can be calculated from the actual distribution via
\begin{align}
    \mu = \sum_y y \mP(y\mid\bx)\,,\quad\sigma^2 = \sum_y (y-\mu)^2 \mP(y\mid\bx)\,.
\end{align}
The idea behind this assumption of normal distributions is, that the $L_{exp}$ should not include any shape details of the true label distribution as they are already included in the $L_{ld}$ component.
Thus, our expectation regression KL loss is
\begin{align}
    L^\ast_{exp} &= \KLD(\mathcal{N}(\cdot\mid\mu,\sigma^2)||\mathcal{N}(\cdot\mid\hat{\mu},\hat{\sigma}^2)) = \overset{\ref{kl_gauss}}{\dots} = \log\frac{\hat{\sigma}}{\sigma} + \frac{\sigma^2 + (\hat{\mu} - \mu)^2}{2\hat{\sigma}^2} - \frac{1}{2}\,.
\end{align}
A detailed derivation of this expression can be found in the Appendix~\ref{kl_gauss}.
Furthermore, we want to improve convergence, if the model predicts \enquote{spiky} distributions, especially in early stages. Thus, we add a loss that penalizes two neighboring neurons having large differences. In other words: We impose a smoothness constraint on the learned distribution by adding a new loss component.

Since each neuron represents a point on a probability mass function, we can also write $\hat{\mP}(y_i\mid\bx) = {\hat{\mP}_i}$, where $\hat{\mP}_i$ is the contribution to the pmf prediction of the $i$-th neuron in the dense output layer. To formulate such a \enquote{smoothness} criterion, we calculate the KL divergence between the predicted pmf and itself, shifted by one position. Since the KL divergence is not commutative, we symmetrize the shift by taking the average of both shift directions. Thus,
\begin{align}
    L^\ast_{smooth} &= \frac{1}{2} \left[ \KLD(\hat{\mP}||\hat{\mP}^s) + \KLD(\hat{\mP}^s||\hat{\mP})\right] = \overset{\ref{kl_smooth}}{\dots} = \frac{1}{2} \sum_y (\hat{\mP}(y\mid\bx) - \hat{\mP}^s(y\mid\bx)) \log\frac{\hat{\mP}(y\mid\bx)}{\hat{\mP}^s(y\mid\bx)}\,,
\end{align}
by using the KL divergence definition from Equation~\eqref{kl_div}. Therein, $\hat{\mP}^s$ denotes a shifted version of $\hat{\mP}$ with $\hat{\mP}_i^s = \hat{\mP}_{i+1}$. Intermediate steps can again be found in the Appendix~\ref{kl_smooth}. Our novel loss component can be interpreted as a regularization term but, again, without any hyperparameter.
Note, that another advantage of a full KL loss is the scale invariance. All loss components are based on a summation over operations on single pmf components, i.e. neurons. Neither the position itself, nor the distance between neurons matters. This can be important for a multi-dimensional regression where features may have different units or domains, for instance if the model should predict age and head pose angle~\cite{dldl}.

We combine the 3 loss components by a simple sum. Thus, our new full KL loss finally is
\begin{align}
    L^\ast = L_{ldl} + L^\ast_{exp} + L^\ast_{smooth}\,.
\end{align}

\begin{figure}[!htb]
    \centering
    \includegraphics[width=0.8\textwidth]{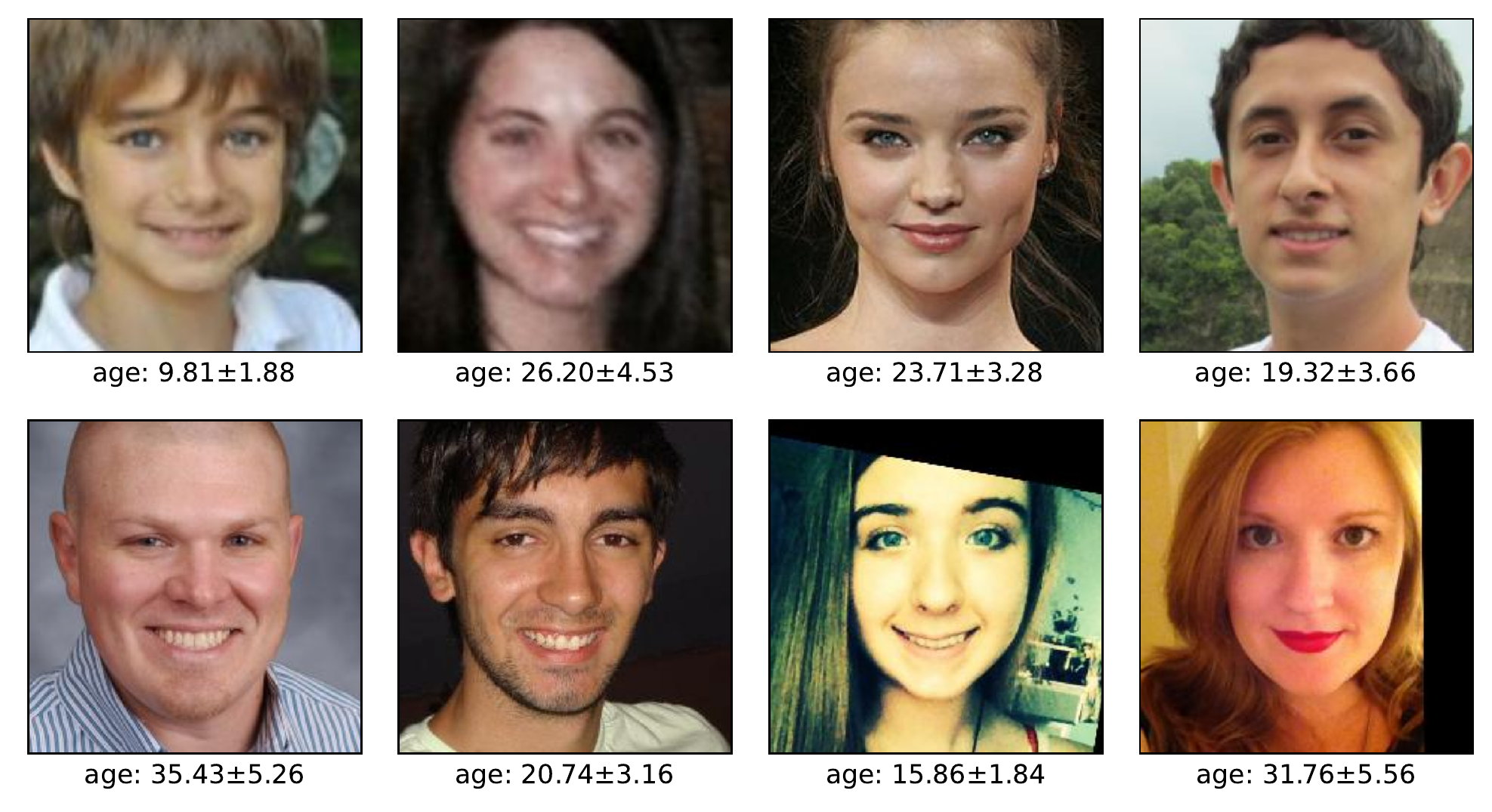}
    \caption{Examplary images $\bx$ of the ChaLearn16 dataset~\cite{chalearn16}. The face positions are aligned for all images and the image size is unified to $\num{224}\times\SI{224}{px}$. Below each image, the mean and std of the age estimation based on many individual guesses are given.}
    \label{fig:chalearn_sample}
\end{figure}

\section{Experimental Results}
We conduct some numerical experiments to evaluate the performance of our novel DLDL loss.

In order to preserve comparability between our results and the results on the original DLDL loss, we do not change the training procedure\footnote{Adam optimizer, parameters: $lr=\num{1e-3}, \beta=(\num{0.9},\num{0.999}), \epsilon=\num{e-8}$; $lr$ is decreased with factor \num{0.1} every 30 epochs; batch size $ = \num{128}$~\cite{dldl}} and choose the \textit{ThinAgeNet} architecture which we pretrained on the MS-Celeb-1M data set~\cite{msceleb} as in the original work~\cite{dldl}. Next, we finetuned the model on the ChaLearn16 data set~\cite{chalearn16} (c.f. Fig.~\ref{fig:chalearn_sample}). The dataset contains aligned faces with age estimations based on individual human annotations. Thus, each image has a normal label distribution with given mean and standard deviation. We conduct 20 runs with 10 different seeds and performed a run with our and the reference loss for each seed. The model covers ages between \num{0} and \num{100} years with a distance of $\Delta l = \num{1}$ year. Thus, the last layer has \num{101} neurons, one for each year. Some example images of the used dataset are shown in Figure~\ref{fig:chalearn_sample}. We do not vary $\Delta l$ as the influence of this parameter is found to be marginal already in~\cite{dldl}.

\begin{figure}[!htb]
    \centering
    \begin{subfigure}{0.48\textwidth}
        \includegraphics[width=\textwidth]{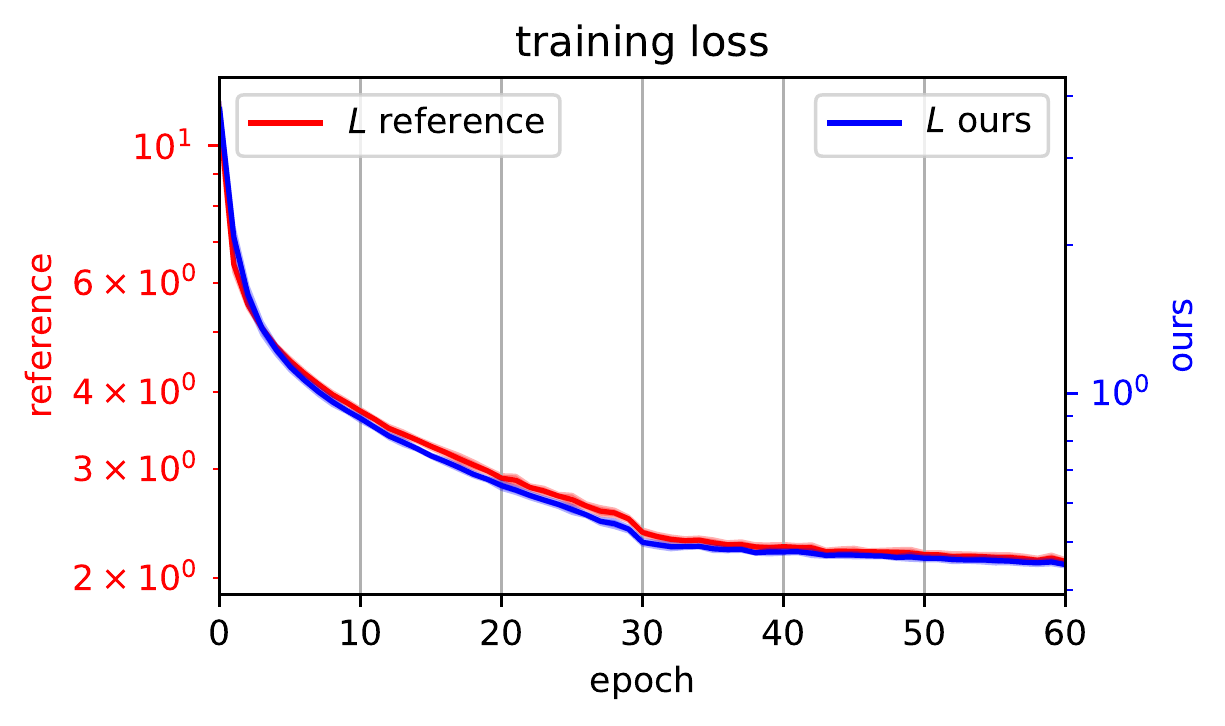}
        \caption{}\label{fig:train_loss}
    \end{subfigure}
    \hfill
    \begin{subfigure}{0.48\textwidth}
        \includegraphics[width=\textwidth]{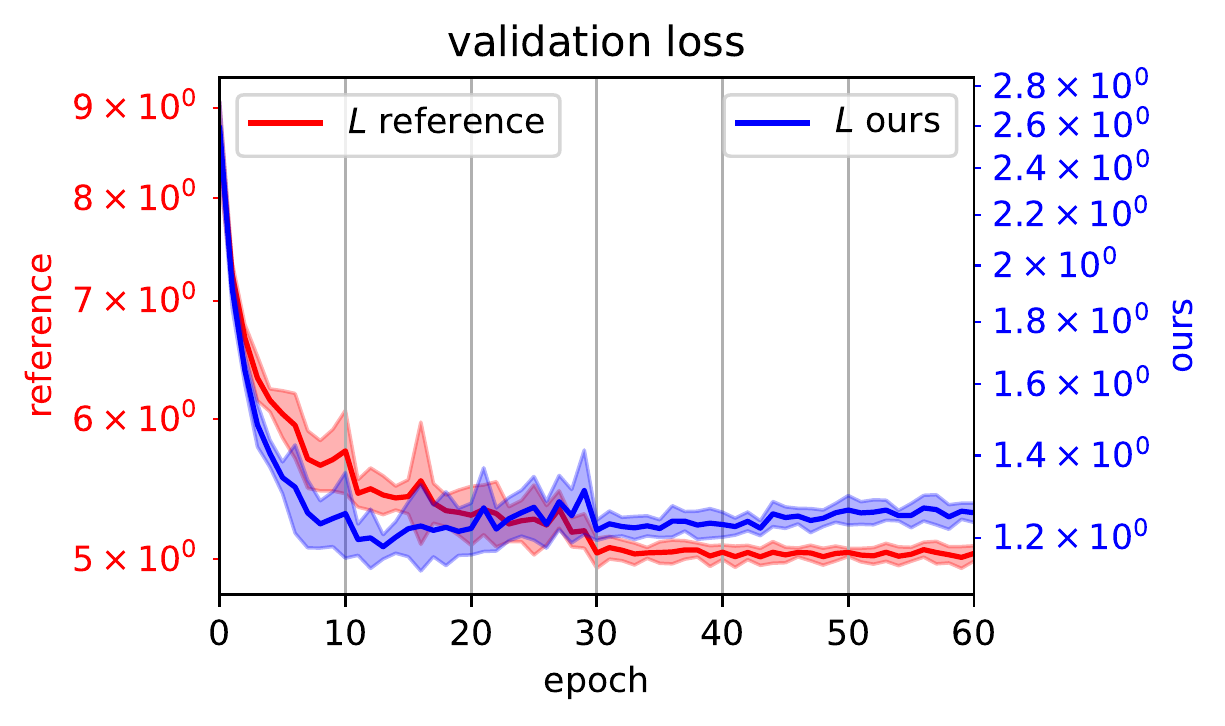}
        \caption{}\label{fig:val_loss}
    \end{subfigure}
    \caption{Training~(\subref{fig:train_loss}) and validation~(\subref{fig:val_loss}) loss curves. Red lines show the reference loss, blue lines show our loss. Mean and standard deviation of the 10 performed runs are denoted by a solid line and a surrounding band, respectively.}
    \label{fig:train_val_loss}
\end{figure}

\begin{figure}[!htb]
    \centering
    \begin{subfigure}{0.48\textwidth}
        \includegraphics[width=\textwidth]{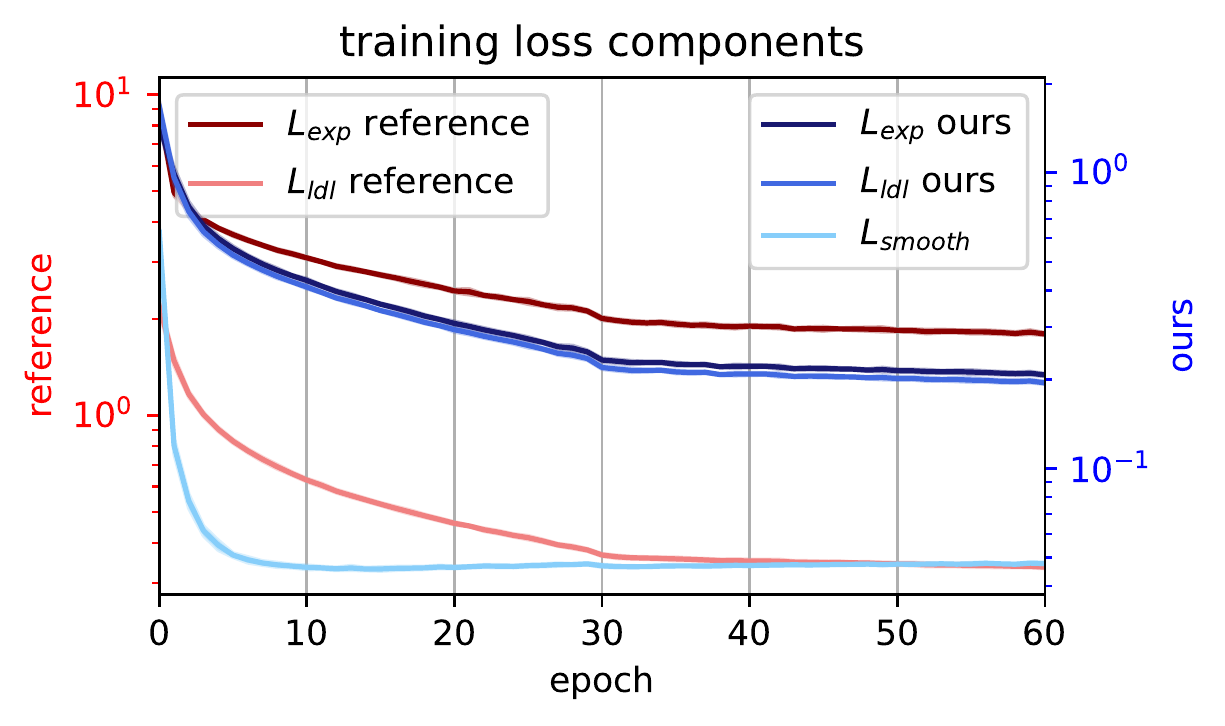}
        \caption{}\label{fig:train_loss_comp}
    \end{subfigure}
    \hfill
    \begin{subfigure}{0.48\textwidth}
        \includegraphics[width=\textwidth]{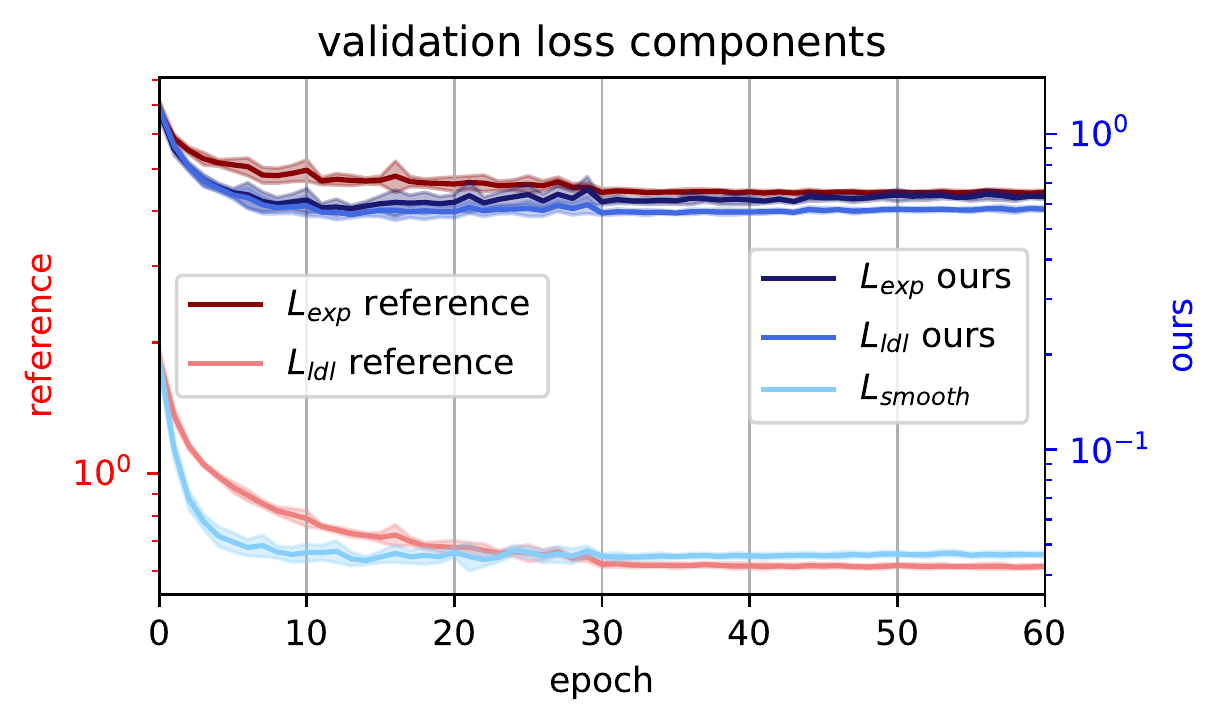}
        \caption{}\label{fig:val_loss_comp}
    \end{subfigure}
    \caption{Components of training~(\subref{fig:train_loss_comp}) and validation~(\subref{fig:val_loss_comp}) loss. The reddish-colored lines show the reference loss components, the blueish-colored lines show our loss components. Mean and standard deviation of the 10 performed runs are denoted by a solid line and a surrounding band, respectively.}
	\label{fig:train_val_loss_comp}
\end{figure}

\begin{figure}[!htb]
    \centering
    \begin{subfigure}{0.48\textwidth}
        \includegraphics[width=\textwidth]{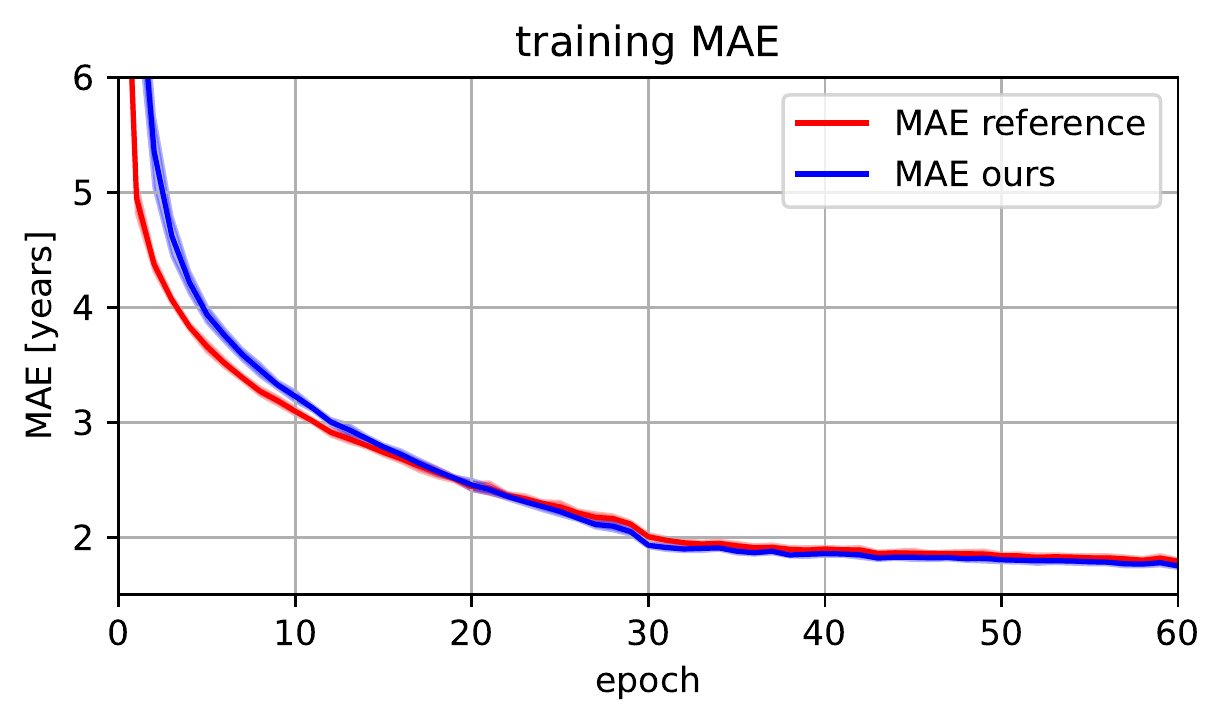}
        \caption{}\label{fig:train_mae}
    \end{subfigure}
    \hfill
    \begin{subfigure}{0.48\textwidth}
        \includegraphics[width=\textwidth]{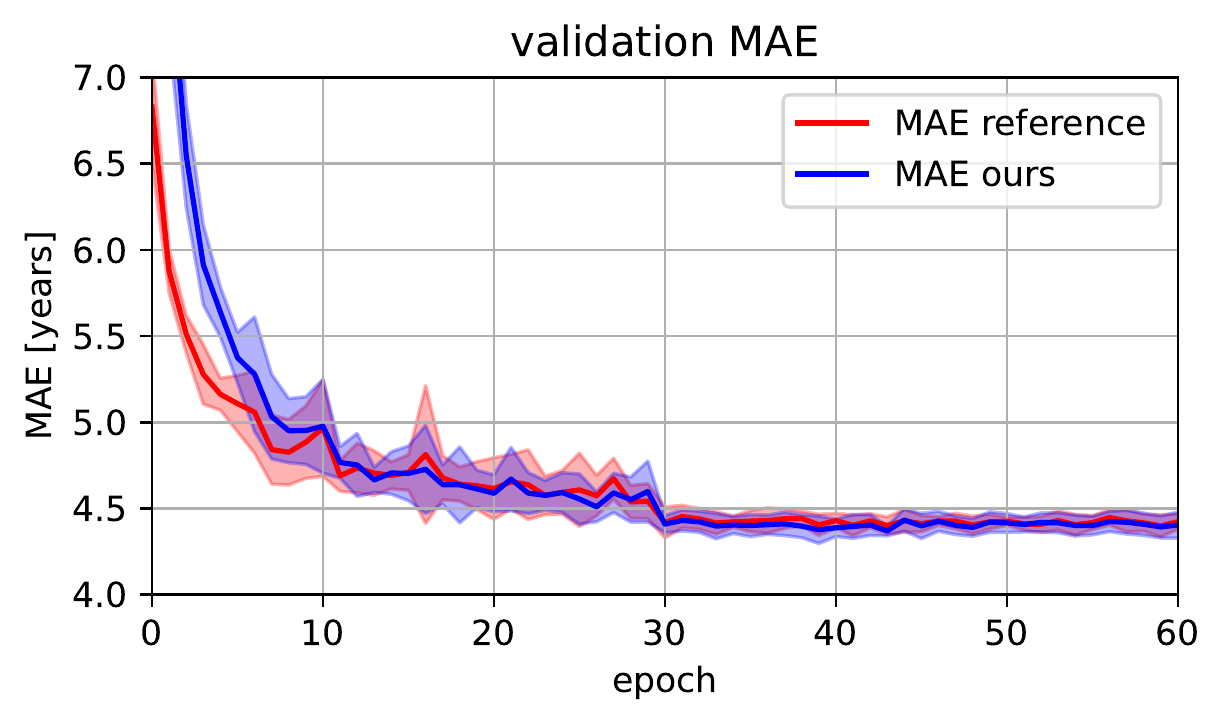}
        \caption{}\label{fig:val_mae}
    \end{subfigure}
    \caption{Mean absolute error (MAE) for training~(\subref{fig:train_mae}) and validation~(\subref{fig:val_mae}) set. Red lines show the reference method, blue lines show our method. Mean and standard deviation of the 10 performed runs are denoted by a solid line and a surrounding band, respectively.}
    \label{fig:train_val_mae}
\end{figure}

In the following, we show the training and validation process for the first \num{60} epochs. The 10 runs each are represented by their mean values and the corresponding standard deviation for all loss terms. We show the reference and loss terms in the same plots with separate y-axes. The left y-axes correspond to the reference loss scale, whereas the right ones correspond to our loss scale 

First of all, it can be stated that all runs were very similar, both for our method and for the reference method. Figure~\ref{fig:train_loss} shows that both losses converge nicely, where our loss seems to converge a bit faster. However, this could be a rather spurious finding, since the losses have different scales that are hardly comparable directly. Figure~\ref{fig:val_loss} shows a more remarkable fact. With our loss, we can observe, that the model indeed shows over-fitting after about \num{15} epochs. The reference validation loss, however, keeps decreasing. This gives a hint, that our loss might parameterize the label distribution learning task more accurately. As for our method the loss components are directly comparable to each other, we can take a look on the single validation loss components to observe where the over-fitting exactly happens. Figure~\ref{fig:val_loss_comp} shows the validation loss curves split into the single components. At first, we see that the label distribution and expectation regression loss have indeed the same scale in our loss, whereas in the conventional loss, the orders of magnitude are different per se and have to be tuned by a hyperparameter, if necessary.

The smoothness penalty loss is only important in early training stages as it drops to an order of magnitude below the other components in the later stage. However, we notice, that the over-fitting is mainly caused in the expectation regression loss since we see a slight increase in our $L_{exp}$ component after about 15 epochs. A look at the mean absolute error (MAE) of the age estimation at Figure~\ref{fig:train_val_mae} shows, that both reference and our method converge to very similar results, although the reference method converges slightly faster at early stages. A reason for that could be, that the MAE is proportional to the L1 loss and, thus, is directly optimized by the reference method. In our method, the MAE is optimized indirectly. An interesting observation is, that the over-fitting suggested by our validation loss curve can not be found in the MAE. Thus, the over-fitting might happen regarding standard deviations of the label distribution, since the MAE does not represent the label distribution but only the expectation value.

\section{Discussion}
We proposed a novel DLDL loss that simplifies problem modeling and potentially associated parameter tuning by removing a hyperparameter. Secondly, we see, that our method is more sensible to the actual label distribution. This is an advantage if the model should, besides the expectation value, also give a precise uncertainty estimation. Thirdly, the scale invariance of our loss formulation is a major advantage if a regression is performed on several dimensions with diverse scales and units, simultaneously.

Future research may concern an evaluation of our novel loss on some multi-dimensionally labeled data. In such a case, one might benefit even more from the unified scale of our loss components.
\begin{acknowledgments}
    Parts of this work have been funded by the Deutsche Forschungsgemeinschaft (DFG, German Research Foundation) under Germany’s Excellence Strategy – EXC 2070 – 390732324 and partially by the European Agriculture Fund for Rural Development with contribution from North-Rhine Westphalia (17-02.12.01 - 10/16 – EP-0004617925-19-001). Other parts have been funded by the Federal Ministry of Education and Research of Germany as part of the competence center for machine learning ML2R (01S18038B).
\end{acknowledgments}

\newpage
\bibliography{main}

\appendix

\section{Appendix}

\subsection{KL divergence between two univariate, discrete normal distributions}\label{kl_gauss}
\begin{align*}
    & \KLD(\mathcal{N}(\cdot\mid\mu,\sigma^2)||\mathcal{N}(\cdot\mid\hat{\mu},\hat{\sigma}^2))\\
    &= \sum_y \mP(y\mid\bx) \log \mP(y\mid\bx) - \sum_y \mP(y\mid\bx) \log\hat{\mP}(y\mid\bx)\\
    &= -\sum_y \mP(y\mid\bx)\left[\frac{1}{2}\log(2\pi\sigma^2) + \frac{(y-\mu)^2}{2\sigma^2}\right]
    + \sum_y \mP(y\mid\bx)\left[\frac{1}{2}\log(2\pi\hat{\sigma}^2) + \frac{(y-\hat{\mu})^2}{2\hat{\sigma}^2}\right]\\
    &= -\frac{1}{2}\log(2\pi\sigma^2)\underbracket{\sum_y \mP(y\mid\bx)}_{=1} - \frac{1}{2\sigma^2}\underbracket{\sum_y (y-\mu)^2 \mP(y\mid\bx)}_{=\sigma^2}\\
    & \quad + \frac{1}{2}\log(2\pi\hat{\sigma}^2)\underbracket{\sum_y \mP(y\mid\bx)}_{=1} + \frac{1}{2\hat{\sigma}^2}\sum_y (y-\hat{\mu})^2 \mP(y\mid\bx)\\
    &= \log\frac{\hat{\sigma}}{\sigma} - \frac{1}{2} + \frac{\overbracket{\sum_y y^2 \mP(y\mid\bx)}^{=\mu^2 + \sigma^2} - 2\hat{\mu}\overbracket{\sum_y y \mP(y\mid\bx)}^{=\mu} + \hat{\mu}^2}{2\hat{\sigma}^2} \\
    &= \log\frac{\hat{\sigma}}{\sigma} - \frac{1}{2} + \frac{\sigma^2 + (\hat{\mu} - \mu)^2}{2\hat{\sigma}^2}
\end{align*}

\subsection{KL-formulated smoothness penalty}\label{kl_smooth}
\begin{align*}
    & \frac{1}{2} \left[ \KLD(\hat{\mP}||\hat{\mP}^s) + \KLD(\hat{\mP}^s||\hat{\mP})\right]\\
    &= \frac{1}{2} \left[ \sum_y \hat{\mP}(y\mid\bx)\log\hat{\mP}(y\mid\bx) - \sum_y \hat{\mP}(y\mid\bx)\log\hat{\mP}^s(y\mid\bx) \right. \\
    & \quad + \left. \sum_y \hat{\mP}^s(y\mid\bx)\log\hat{\mP}^s(y\mid\bx) - \sum_y \hat{\mP}^s(y\mid\bx)\log\hat{\mP}(y\mid\bx) \right]\\
    &= \frac{1}{2} \left[ \sum_y \left(\hat{\mP}(y\mid\bx)-\hat{\mP}^s(y\mid\bx)\right)\log\hat{\mP}(y\mid\bx) - \left(\hat{\mP}(y\mid\bx)-\hat{\mP}^s(y\mid\bx)\right)\log\hat{\mP}^s(y\mid\bx) \right]\\
    &= \frac{1}{2} \sum_y (\hat{\mP}(y\mid\bx) - \hat{\mP}^s(y\mid\bx)) \log\frac{\hat{\mP}(y\mid\bx)}{\hat{\mP}^s(y\mid\bx)}
\end{align*}

\end{document}